\documentclass{article}
\usepackage{ijcai22}

\usepackage{times}
\usepackage{soul}
\usepackage{url}
\usepackage[hidelinks]{hyperref}
\usepackage[utf8]{inputenc}
\usepackage[small]{caption}
\usepackage{graphicx}
\usepackage{amsmath}
\usepackage{amsthm}
\usepackage{booktabs}
\usepackage{algorithm}
\usepackage{algorithmic}
\DeclareMathOperator*{\argmax}{arg\,max}

\usepackage{bm}
\usepackage{xcolor}
\usepackage{color}
\usepackage{graphicx}
\usepackage{wrapfig}

\newcommand{\remove}[1]{{}}

\usepackage{caption}
\usepackage{subcaption}
\usepackage{multirow}
\graphicspath{{figures/}}

\title{Network Graph Based Neural Architecture Search}

\author{
Zhenhan Huang$^1$\and
Chunheng Jiang$^1$\and
Pin-Yu Chen$^2$\And
Jianxi Gao$^1$\footnote{Contact Author}\\
\affiliations
$^1$Department of Computer Science, Rensselaer Polytechnic Institute, Troy, NY 12180\\
$^2$IBM Thomas J. Watson Research Center, Yorktown Heights, NY 10598\\
\emails
\{huangz12, jiangc4\}@rpi.edu,
pin-yu.chen@ibm.com,
gaoj8@rpi.edu
}

\begin{document}

\maketitle

\begin{abstract}
  Neural architecture search enables automation of architecture design. Despite its success, it is computationally costly and does not provide insight on how to design a desirable architecture. Here we propose a new way of searching neural network where we search neural architecture by rewiring the corresponding graph and predict the architecture performance by graph properties. Because we do not perform machine learning over the entire graph space and use predicted architecture performance to search architecture, the searching process is remarkably efficient. We find graph based search can give a reasonably good prediction of desirable architecture. In addition, we find graph properties that are effective to predict architecture performance. Our work proposes a new way of searching neural architecture and provides insights on neural architecture design.
\end{abstract}

\section{Introduction}

Neural networks architecture achieves a great success over last few years in various challenging applications, such as image classification \cite{krizhevsky2012imagenet}, speech recognition \cite{hinton2012deep} and machine translation \cite{wu2016google}. The success, largely attributed to the feature engineering, is accompanied by the arise of architecture engineering, e.g., AlexNet \cite{krizhevsky2012imagenet}, VGGNet \cite{simonyan2014very}, ResNet \cite{he2016deep}. Increasingly complicated neural network architectures make it progressively more difficult for manual design of architecture. Neural architecture search (NAS), enabling automation of architecture engineering, proves its potential in boosting the performance of neural network architecture. For example, NAS methods have outperformed some manually designed architecture in some applications, such as image classification \cite{real2019regularized}, object detection \cite{zoph2018learning}.

In NAS method, there are three major components \cite{elsken2019neural}: search space $\mathcal S$, search strategy ${\mathcal A}_{s}$ and performance estimation strategy ${\mathcal A}_{e}$. Predefined $\mathcal S$ confines the total number of possible architectures if it is not unbounded. Thus it will affect the search efficiency and optimal architecture. ${\mathcal A}_{s}$ determines the search efficiency and should avoid potential pitfall of local minimum. ${\mathcal A}_{e}$ provides a way to evaluate architecture candidate and feedback. The simplest way for ${\mathcal A}_{e}$ is to perform a standard training and validation for targeting optimal architecture.

Powerful as NAS method is, searching an architecture in $\mathcal S$ and testing its performance will take ample time. Although decomposing hand-crafted architectures into motifs and searching motifs, blocks or cells can boost searching speed, this process inevitably introduces bias in the search space. Furthermore, NAS method does not throw light on why specific architectures outperform others and general principles of designing an architecture. Network graphs, on the other hand, have good metrics for evaluation. For example, clustering coefficient measures the degree to which nodes a in graph cluster. If we can bridge graph and neural network, we will be able to relate graph properties to architecture performance. Instead of searching in architecture space, we can perform search in graph space. By targeting the optimal graph structure, we will able to locate the optimal neural network architecture.

\begin{figure*}[hbt]
    \centering
    \includegraphics[width=1.6\columnwidth]{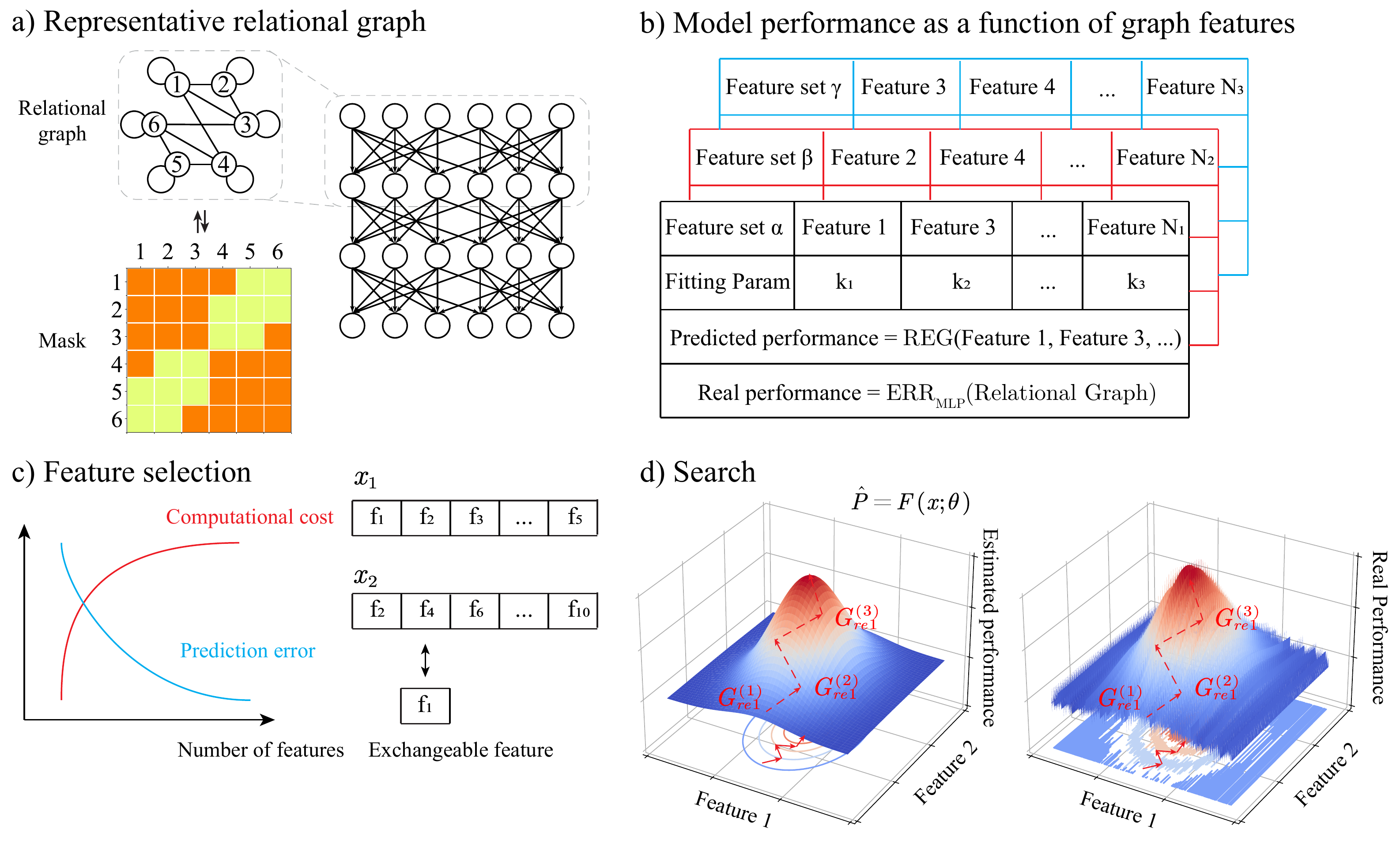}
    \caption{Schematic illustration of graph-based neural architecture search}
    \label{fig:fig_abs}
\end{figure*}

Figure \ref{fig:fig_abs} shows the graph-based neural architecture search. The relationship between graph and neural network is bridged by relational graph. We predict machine learning error based on graph properties. A set of graph features is selected for the prediction. Based on predicted error, we apply rewiring strategy to search a better or worse neural architecture.

\section{Bridge of Neural Network and Graph}

\subsection{Graph Representation of Neural Network}

The relation between graph and neural network can be bridged by relational graph \cite{you2020graph}. For a fixed-width multilayer perceptron (MLP) where each layer contains the same number $R$ of computation units (neurons), suppose the input and output of the $r$-th layer ($1 \le r \le R$) is $\mathbf{X}^{(r)}$ and $\mathbf{X}^{(r + 1)}$, respectively. A neuron in the $r$-th layer computes \cite{you2020graph}:

\begin{equation}
    x^{(r + 1)}_{i} = \sigma (\sum_{j \in N(i)} \omega^{(r)}_{ij}x^{(r)}_{j}),
\end{equation}
where $w^{(r)}_{ij}$ is the $i$-th row and $j$-th column of the trainable weight $\mathbf{W}^{(r)}$, $x^{(r)}_{j}$ is the $j$-th dimension of the input $\mathbf{X}_{j}$, $x^{(r)}_i$ is the $i$-th dimension of the output $\mathbf{X}_{i}$. $\sigma(\cdot)$ is the activation function introducing non-linearity. $N(i)$ is the set of neurons of the $(r + 1)$-th layer connecting to $r$-th layer and defined by relational graph $G$.

\subsection{Undirected Graph Generator}

Graphs can be categorized into two groups: directed graph or undirected graph. We consider relational graph as undirected, so its adjacency matrix is symmetric. The classic graph generation algorithms include (1) Watts-Strogatz (WS) model that can generate graphs with small-world properties \cite{watts1998collective}. The "small world effect" is generally referred to describe graphs whose average path length is comparable with a homogeneous random graphs \cite{prettejohn2011methods}; (2) Erd\H{o}s-R\'{e}nyi (ER) model that can generate random graphs \cite{erdos1960evolution}; (3) Barab\'{a}si-Albert (BA) model that constructs scale-free graphs \cite{barabasi1999emergence}. A scale-free graph has a degree distribution following power law, i.e., the probability of a node having a degree $k$ has a scale-invariant decay $P(k) \sim k^{-\gamma}$, where $\gamma$ is a constant and confined by $\gamma > 1$; (4) Harary model that generates graphs with maximum connectivity \cite{harary1962maximum}. WS-flex graph generator, proposed in \cite{you2020graph}, is a more general form of WS model by relaxing the constraint that all the nodes have the same degree before random rewiring. After fixing the number of nodes to be 64, WS-flex graph generators are capable of generating graphs encompassing almost all graphs constructed by classic graph generation algorithms mentioned above \cite{you2020graph}. 

\section{Experimental Setup}

We use the CIFAR-10 dataset \cite{krizhevsky2009learning} for training MLPs. CIFAR-10 dataset has 50K training images and 10K validation images. Each image is a matrix with a dimension of $32 \times 32 \times 3$. 60K images are classified into 10 categories for supervised machine learning.

\subsection{Graph Space}

We use WS-flex graph generator to generate undirected graphs. The number of nodes is fixed to be $64$ and the average degree range is $[2, 63]$. Total number of generated graphs is $5983$. We calculate $26$ Graph properties including average degree, clustering coefficient, heterogeneity, average path length, bimodularity, greedy modularity, resilience parameter, degree entropy, wedge count, gini index, average node connectivity, edge connectivity, average closeness centrality, average closeness centrality (WF improvement), average eccentricity, diameter, radius, average edge betweenness centrality, average node betweenness centrality, central point of dominance, core number, minimum Laplacian spectrum, maximum Laplacian spectrum, transitivity, local efficiency, global efficiency. Detailed description regarding to those properties can be referred in Appendix \ref{appendix:all_features}.

We find that heterogeneity range for $5983$ graphs constructed by ws-flex graph generator is $[0, 0.82]$. Most graphs are not heterogeneous in the structure. To introduce more heterogeneous graphs, we do random rewiring based on those graphs. Total number of graphs becomes $19724$.

\subsection{Feature Selection}

Average path length and clustering coefficient are considered as good indicator for predicting learning error of MLP \cite{you2020graph}. We calculate 28 features for each graph and take into consideration the potential of these features for predicting machine learning performance. To filter features, we use sequential forward selection (SFS) algorithm \cite{marcano2010feature,ververidis2005sequential,cotter1999forward}, a commonly used method for reducing the data dimension. SFS algorithm is a bottom-up search strategy in which an empty set $S$, the starting point, continuously add features until all features are added. At each iteration, $S$ greedily searches for best feature from remaining feature pool and include it.

Selecting MLP learning error as output variable and graph properties as input variables, We use linear regression to fit results and choose mean squared error (MSE) as metric to select features. We use random splitting strategy to obtain training set and test set. The ratio of training set to test set is $9:1$. At each iteration of SFS algorithm, we use training set to determine linear regression parameters and test set to calculate evaluation parameters MSE and Pearson correlation coefficient.

\subsection{Neural Network Architecture}

We use a 5-layer MLPs as the neural network architecture. Each MLP layer has 512 hidden units as baseline architecture. The relational graph determines the connection of hidden units between two neighboring layer. To ensure all networks have the approximately same complexity, we use FLOPS (\# of multiplication and addition) as metric to adjust baseline architecture such that FLOPS for different relational graph represented networks is roughly the same.

Fig. \ref{fig:fig_abs} a) shows the illustration of MLP architecture. Each MLP layer contains Batchnorm ReLU layer to introduce non-linearity and BatchNorm layer \cite{ioffe2015batch}. Batch size in the training process is 128 and total number of epochs for training a model is 200. We use a decaying learning rate with a cosine annealing \cite{loshchilov2016sgdr}. Our Momentum Optimizer has an initial learning rate of 0.1, 5e-4 weight decay, 0.9 momentum and uses Nesterov Momentum \cite{sutskever2013importance,nesterov1983method}.

\subsection{Rewiring Algorithm}
To search the best neural architecture, 
we adopt a greedy strategy 
which is different from the conventional approaches in NAS.
We start from a neural architecture with known performance, 
improving the architecture by iteratively rewiring it.
Let $G_t$ be the current architecture, 
$P(G_t)$ be the post-training performance of $G_t$,
and $\mathcal A$ be our rewiring strategy, 
we can formulate the rewiring procedure as

\begin{equation}\label{eq:rewiring}
G_{t+1} = \argmax_{\mathcal A} P(\mathcal A(G_t)).
\end{equation}

Nevertheless, training each $\mathcal A(G_t)$ for the performance $P(\mathcal A(G_t))$ 
will make the search inevitably expensive. 
To address this issue, we refer to a surrogate parametric model $F(G;\bm\theta)$ for $P(G)$, 
which is designed to capture the relationship between 
a set of simple topological properties (see Appendix \ref{appendix:all_features}) and the performance of $G_t$,
i.e. $\hat P(G)=F(G;\bm\theta)$. 
To some extend, the ``goodness" of $F$ determines our choices of the resultant architectures via rewiring. 
We search the architecture by selecting a good rewired candidate or discarding a bad one. 
If $F(G_t;\bm\theta)$ is consistent with the real performance $P(G_t)$,
there will be very few mistakes in our choices and 
our search will guarantee a constantly improved architecture.
Also, some architectures may be a local optimal, 
which can not be rewired to reach a better one.


As shown in Algorithm \ref{alg:rewire_n_search}, 
our approach starts from a randomly selected relation graph 
whose associated neural network post-trained performance may be arbitrarily low. 
We perform a greedy search of a sequence of rewiring operations to
constantly improve the performance of the associated neural network of the initial relational graph. 
Currently, our rewiring strategy allows removal of edges $\mathcal A_{\rm rmv}$, 
building new edges $\mathcal A_{\rm new}$, random rewiring $\mathcal A_{\rm rnd}$ 
and double edge swaps $\mathcal A_{\rm swap}$.

\begin{algorithm}[ht]
\caption{Rewiring and Searching}
\label{alg:rewire_n_search}
\textbf{Input}: A randomly selected relation graph $G_0\in \mathcal G$, 
four rewiring operations $\{\mathcal A_{\rm new},\mathcal A_{\rm rmv},\mathcal A_{\rm swap},\mathcal A_{\rm rnd}\}$, accepted relative improvement $\varepsilon\in (0,1)$, maximum number $K$ of rewiring operations, the performance predictor $F(G;\bm\theta)$ 
of the associated neural network $\phi(G)$ with the topological properties of a relational graph $G$\\
\textbf{Output}: Rewired relation graph $G_K$
\vskip1pt
\begin{algorithmic}[1]
\STATE Let $k=0$, $\hat P_0=F(G_0;\bm\theta)$.
\WHILE{$k<K$}
\STATE Randomly select a rewiring operation $\mathcal A$
\STATE Perform $\mathcal A$ over $G_{k-1}$ and obtain $\hat G_{k-1}=\mathcal A(G_{k-1})$
\STATE Predict the post-trained performance of $\phi(\hat G_{k-1})$ 
with $\hat P_k=F(\hat G_{k-1};\bm\theta)$
\IF {$|\hat P_k/\hat P_{k-1}-1|\ge \varepsilon$}
\STATE Accept $\mathcal A$ with $G_k=\hat G_{k-1}$ and set $k=k+1$
\ELSE
\STATE Reject $\mathcal A$ and repeat the above procedure
\ENDIF
\ENDWHILE
\end{algorithmic}
\end{algorithm}

\section{Experimental Result}

\subsection{Features for prediction}

Calculating all features of a graph can be computational expensive. Hence, we are interested in how many features are needed to predict MLP performance and what are important features. We use SFS algorithm to quantitatively show the relation between number of features and prediction quality, as well as to determine significant features.

\begin{figure}[hbt]
    \centering
    \includegraphics[width=0.7\linewidth]{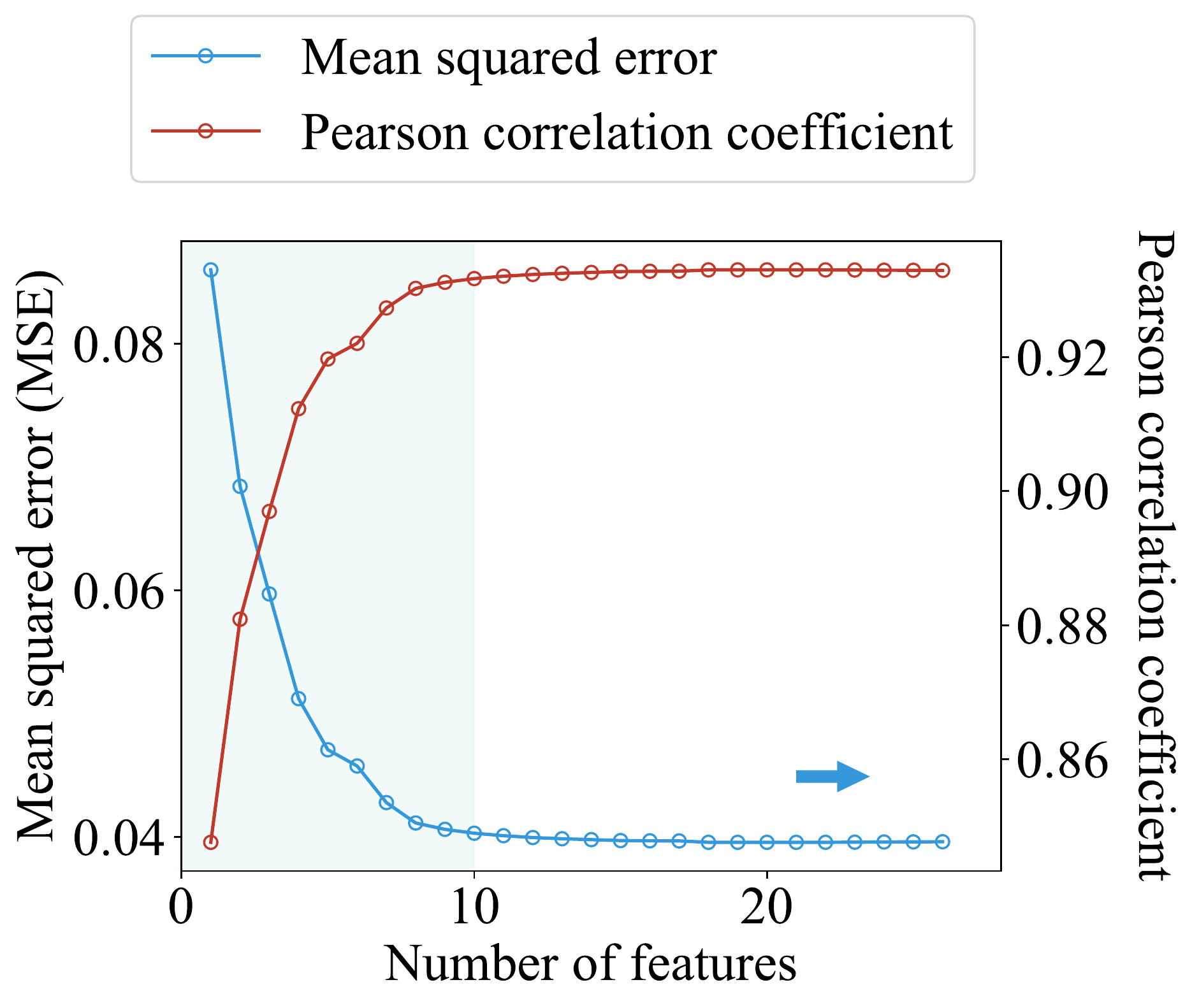}
    \caption{Sequential forward selection for feature selection}
    \label{fig:sfs_select}
\end{figure}

Figure \ref{fig:sfs_select} shows MSE and Pearson correlation coefficient variation with including features. Instead of typical oscillating trend of SFS algorithm, MSE shows a monotone decreasing with adding features. Even though more features give a better prediction quality, the gain in prediction boost becomes negligible while the computational cost increases dramatically when the number of features is more than $10$. At the same time, we find linear regression can give a good prediction. If we only choose one feature, MSE is $0.086$. Adding just $4$ more features, MSE drops to $55\%$ that of one feature. Considering the trade-off between prediction performance and computation cost, we choose the first 10 features to predict MLP learning error. The corresponding MSE is $0.4$ and we have a high $0.93$ Pearson correlation coefficient. The finding suggests that a small number of features can still give a good prediction of MLP performance.

SFS algorithm determines the eccentricity as the starting point. We are interested, within a certain tolerance on prediction error of MLP performance, can we choose other features as starting point that give a similar prediction quality? There are several benefits for moderately sacrificing prediction quality: 1) the computational cost for different features can vary a lot. If we replace features having a high computational cost with ones having low cost and, at the same time, the prediction quality does not change much, the MLP prediction model will be satisfying; 2) there are highly developed theories about some of features. If we use those features for prediction, it is beneficial to understand why and how a neural network has a good performance. For example, the eccentricity measures the maximum distance from one node to all other node. The average path length measures the average of minimum distance between all pairs of nodes. These two features are closely related to message exchange efficiency. If we can replace the eccentricity with the average path length, we will be able to analyze neural network performance using the average path length related theories, e.g., small world network theory; 3) in reality, it is unlikely a certain rewiring strategy just alters one graph property while other property remains unchanged. If we can replace less controllable feature with more controllable one, it becomes easier to target optimal relational graph.

\begin{figure}[hbt]
    \centering
    \includegraphics[width=0.7\linewidth]{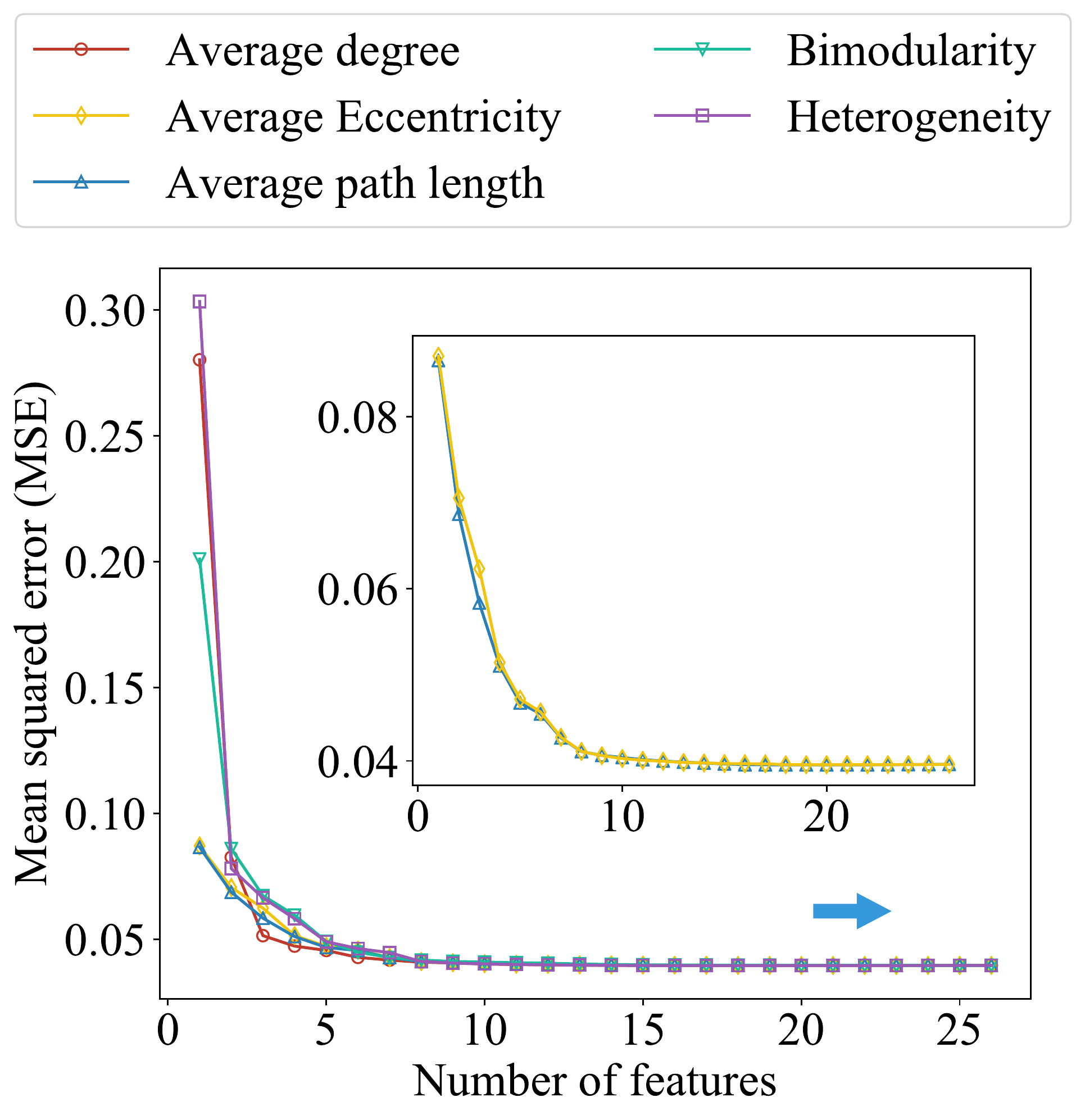}
    \vspace{-4pt}
    \caption{Representative sequential forward selection after fixing the first feature}
    \label{fig:fsfs_select}
\end{figure}

We replace the first feature, i.e., eccentricity, with other features and perform SFS algorithm to determine remaining features. Figure \ref{fig:fsfs_select} shows representative MSE variation as a function of number of features in SFS algorithm after fixing the first feature. Generally, a different starting point of SFS algorithm will lead to different feature set $S$ and different MSE. However, some features give a similar prediction quality and feature set. For example, MSE for average path length and eccentricity at each iteration is very close. Besides, the order of adding features is nearly the same. When we consider the first $10$ features excluding first feature, the eccentricity and the average path length have the same remaining $9$ features. The feature adding order is also the same.

When a single feature is used for predicting MLP performance, calculated features have a pronounced difference in the prediction quality. Some features, such as heterogeneity, have a high $0.3$ MSE. Nevertheless, with adding more features, all features show a fast converge (adding less than 10 features) to a small $0.4$ MSE. This indicates that a linear combination of features gives a good prediction of MLP performance. More details about MSE variation with adding features when fixing the first feature can be referred in Appendix \ref{appendix:feature_fix_plot}.

Table \ref{tab:ft_set_alg} shows first $10$ features determined by SFS algorithm and first $10$ features by SFS algorithm after fixing first feature to be the average path length. Except the first feature, two feature sets have the exactly the same features and linear regression parameters associated with these features have the same sign, i.e., consistent positive correlation or negative correlation. In addition, MSE is close for these two feature sets. Hence, we think the average eccentricity and average path length are interchangeable in terms of predicting MLP performance.

{\footnotesize
\begin{table}[hbt]
\centering
\begin{tabular}{p{1.6cm}p{4cm}p{1cm}}
\toprule
Algorithm & Selected Features & Sign\\
\midrule
\multirow{10}{*}{SFS}
    & Average eccentricity & $+$\\
    & Central point of dominance & $+$\\
    & Resilience Parameter & $+$\\
    & Global efficiency & $-$\\
    & Edge connectivity & $-$\\
    & Wedge count & $-$\\
    & Clustering coefficient & $-$\\
    & Average node connectivity & $+$\\
    & Average closeness centrality & $+$\\
    & Greedy modularity & $-$\\
\midrule
    & Average path length & $+$\\
    & Central point of dominance & $+$\\
    & Resilience Parameter & $+$\\
    & Global efficiency & $-$\\
    SFS fixing    & Edge connectivity & $-$\\
    first feature & Wedge count & $-$\\
    & Clustering coefficient & $-$\\
    & Average node connectivity & $+$\\
    & Average closeness centrality & $+$\\
    & Greedy modularity & $-$\\
\bottomrule
\end{tabular}
\caption{Feature sets determined by SFS algorithms}
\label{tab:ft_set_alg}
\end{table}
}

\begin{figure*}[hbt]
    \centering
    \includegraphics[width=0.8\textwidth]{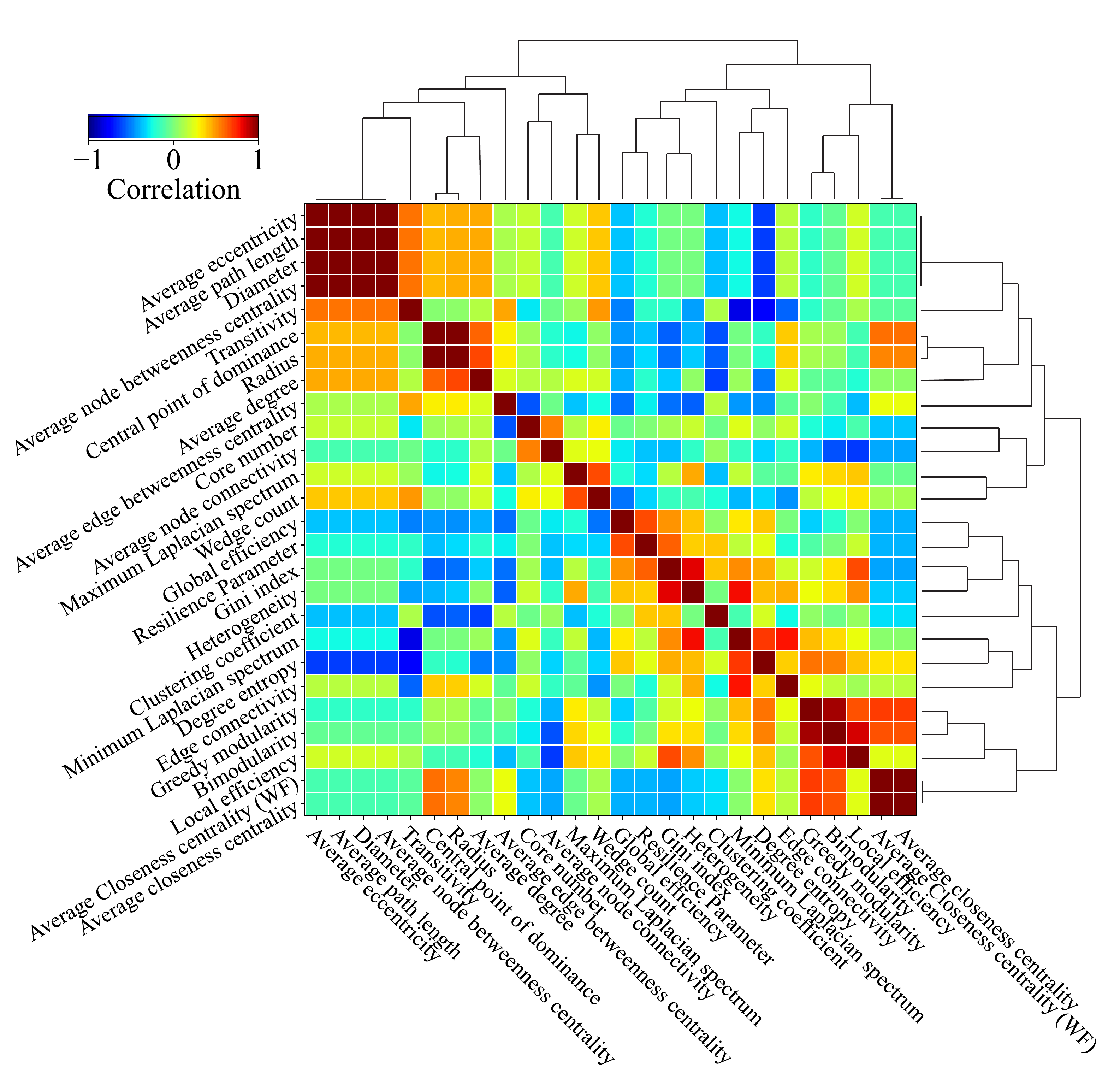}
    \caption{Pearson correlation coefficient of every pair of graph features. Feature sets related to every feature are determined by SFS algorithm}
    \label{fig:ft_heatmap}
\end{figure*}

Except for the pair of the average eccentricity and average path length, we are also curious about a question -- are there other features interchangeable. We use Pearson correlation coefficient to characterize the similarity between every pair of feature sets. We fix first feature and use SFS algorithm to determine the remaining $9$ features. Every remaining $9$ features constitute a set $S$. Then we calculate Pearson correlation coefficient between pairs of those sets. Figure \ref{fig:ft_heatmap} shows Pearson correlation coefficient between all pairs of features. The result reveals the similar effect in predicting MLP performance. The similarity seems to be correlated to physical meaning of graph features. For example, the average path length measures the average of shortest path length between pairs of nodes \cite{albert2002statistical,achard2007efficiency}. The average eccentricity measures the maximum distance from one node to other nodes in a graph $G$ \cite{hage1995eccentricity}. Diameter is the maximum eccentricity of a graph $G$. Average node betweenness centrality calculates the average of node betweenness centrality which is the sum of the fraction of all shortest paths passing through a node \cite{brandes2001faster,brandes2008variants}. These four features are related to message exchange efficiency in a graph. At the same time, they have a highest correlation 1.

\subsection{Prediction of MLP performance}

We choose as the starting point a graph with a highest MLP top 1 error and a graph with lowest MLP top 1 error in the graph pool. Then we use the rewiring strategy described before to rewire graph for searching a better graph structure step by step. At each iteration, if the predicted error decreases by more than a threshold value $0.01$, the rewired graph is accepted and the change of graph structure due to rewiring is kept. To verify the prediction of performance, we use MLPs represented by relational graphs at each stage to calculate the real top 1 error. Total time for rewiring is $6$ hours.

\begin{figure}[hbt]
    \centering
    \begin{subfigure}{0.49\columnwidth}
        \includegraphics[width=\columnwidth]{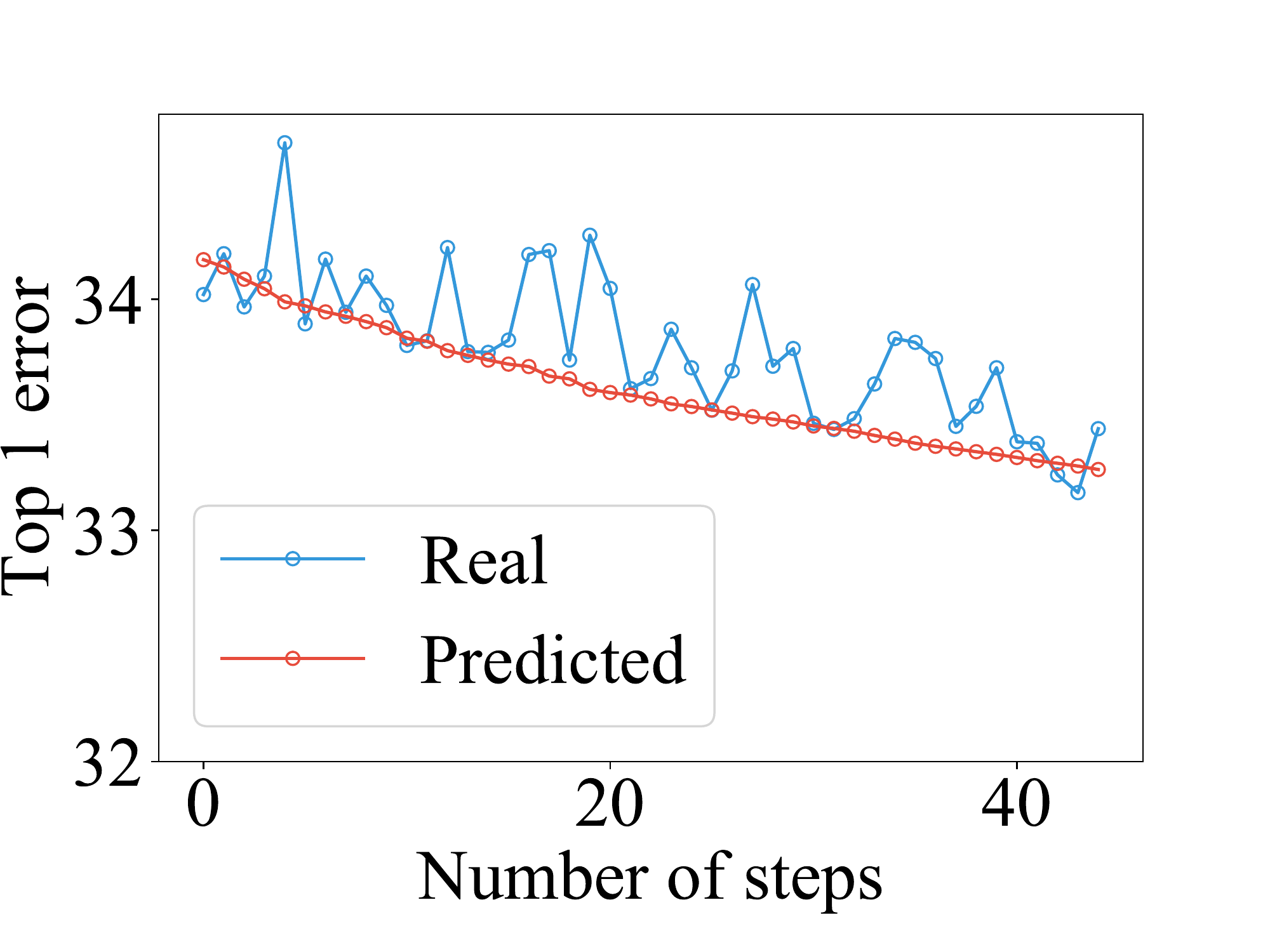}
        \caption{Top 1 error decrease path}
    \end{subfigure}
    \begin{subfigure}{0.49\columnwidth}
        \includegraphics[width=\columnwidth]{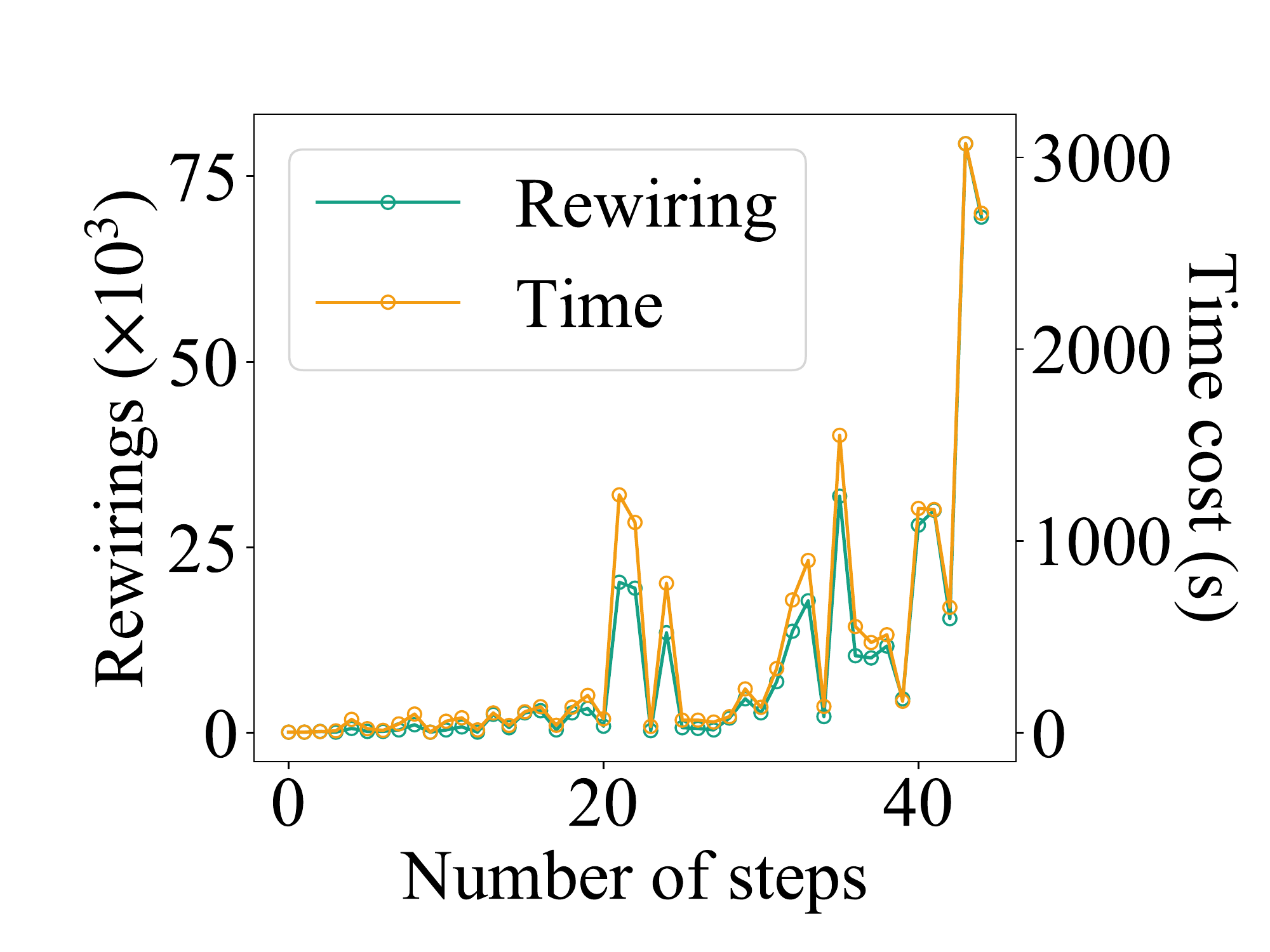}
        \caption{Computational cost}
    \end{subfigure}
    \vspace{-8pt}
    \caption{Rewiring path of MLP learning error decrease}
    \label{fig:mlp_predict_err_decrease}
\end{figure}

\begin{figure}[hbt]
    \centering
    \begin{subfigure}{0.49\columnwidth}
        \includegraphics[width=\columnwidth]{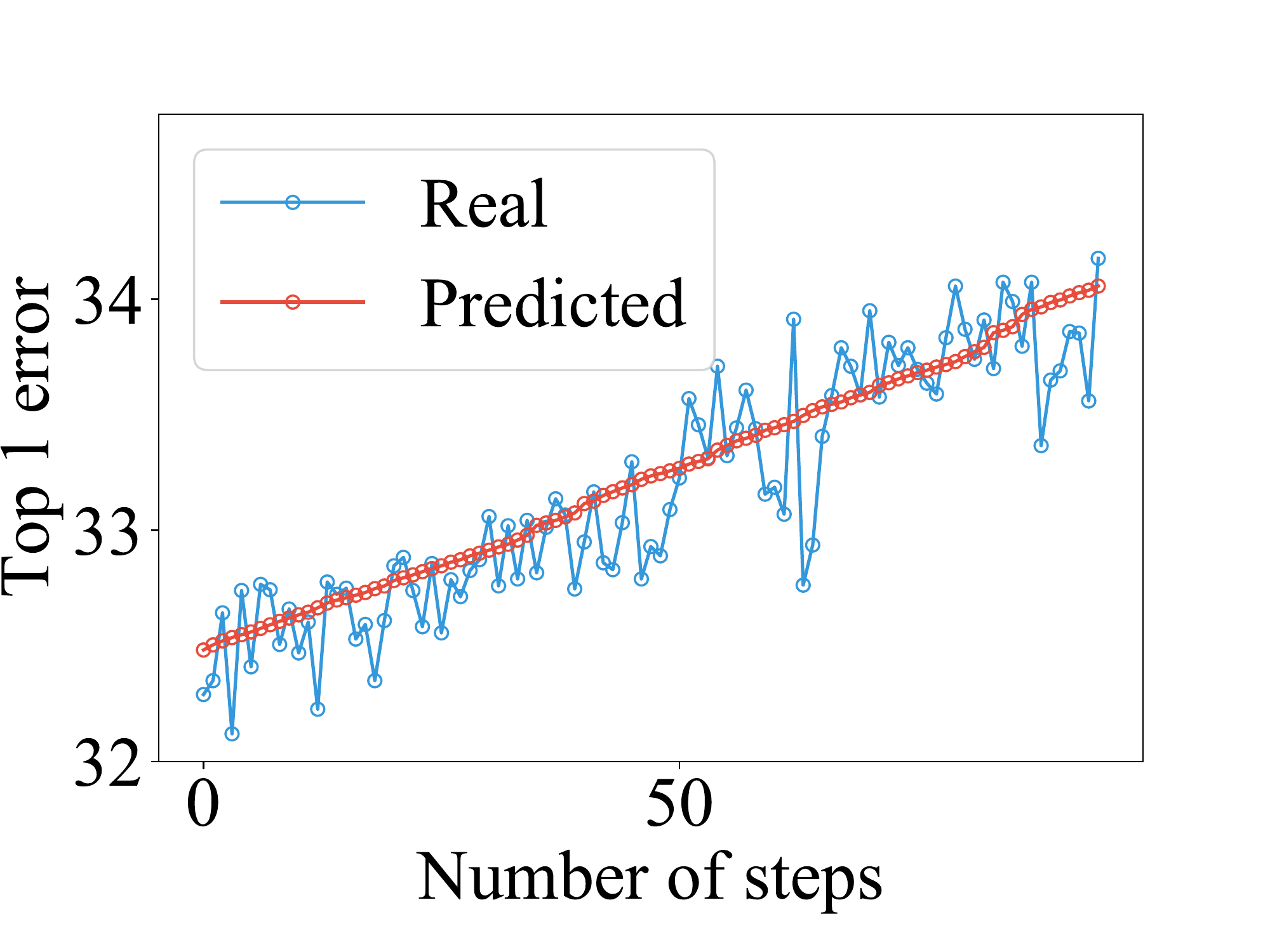}
        \caption{Top 1 error increase path}
    \end{subfigure}
    \begin{subfigure}{0.49\columnwidth}
        \includegraphics[width=\columnwidth]{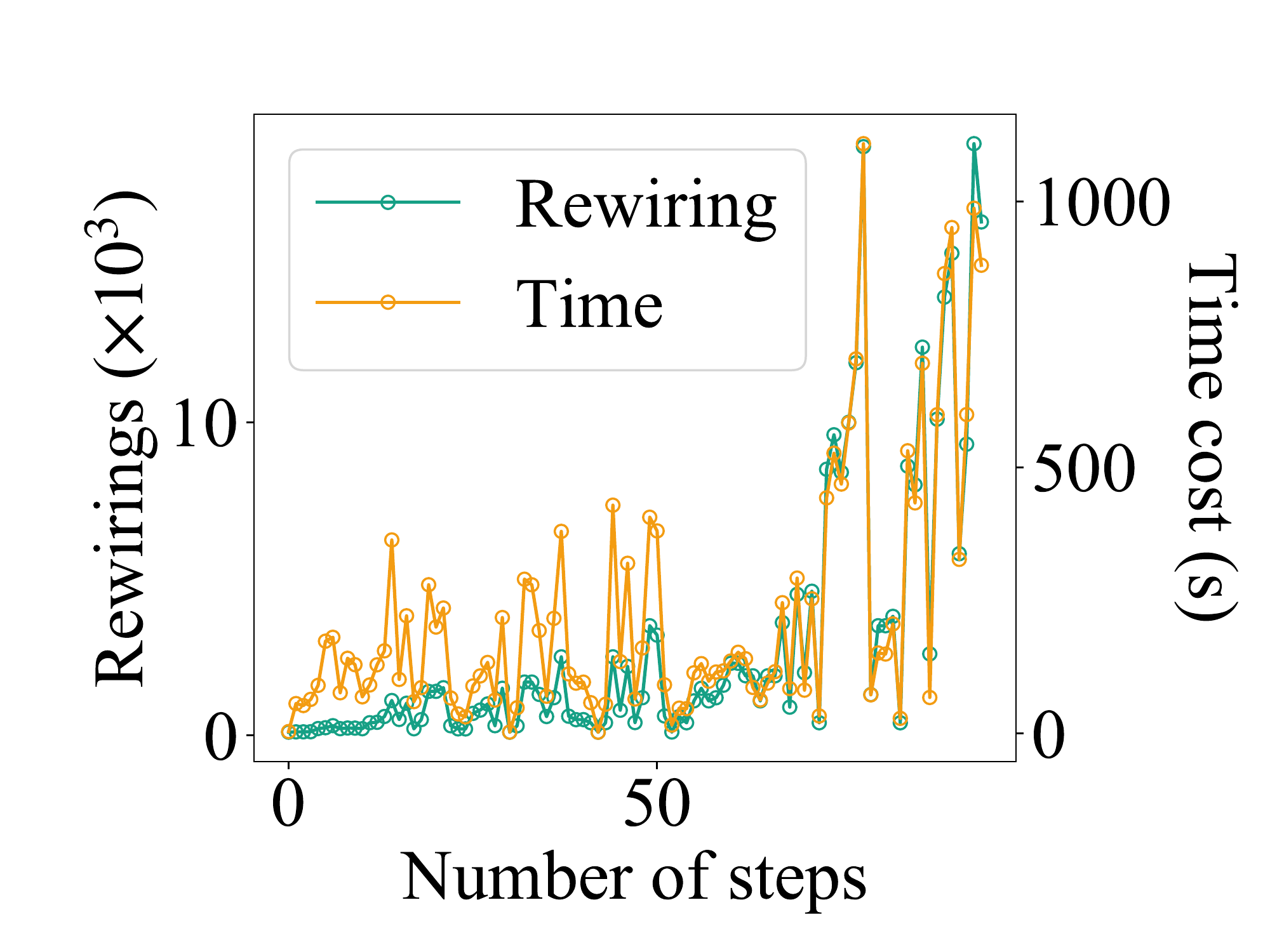}
        \caption{Computational cost}
    \end{subfigure}
    \vspace{-8pt}
    \caption{Rewiring path of MLP learning error decrease}
    \label{fig:mlp_predict_err_increase}
\end{figure}

Figure \ref{fig:mlp_predict_err_decrease} shows the top 1 error decrease path (superior rewiring) and Figure \ref{fig:mlp_predict_err_increase} shows the top 1 error increase path (inferior rewiring). At the end of the rewiring process, both superior rewiring and inferior rewiring show a dramatic increase in the computational cost. In both rewirings, real top 1 error oscillates around the predicted top 1 error. The prediction is promising as the real top 1 error overall follows the prediction path. During the rewiring process, we do not observe the sudden jump in the top 1 error. We think that limited change in the graph edge set does not change MLP performance dramatically. This is consistent with the result from conventaional NAS: small perturbations in the neural network architecture does not have a significant effect on its performance \cite{ru2020neural}. During the superior rewiring process, the computational cost increases drastically even though the learning error does not hit the minimal error region. This might be attributed to premature convergence to the region of suboptimal structure. There seems to be a exploration-exploitation trade-off problem. The greedy searching strategy might lead to a suboptimal graph structure.

\begin{figure}[hbt]
    \centering
    \includegraphics[width=0.7\columnwidth]{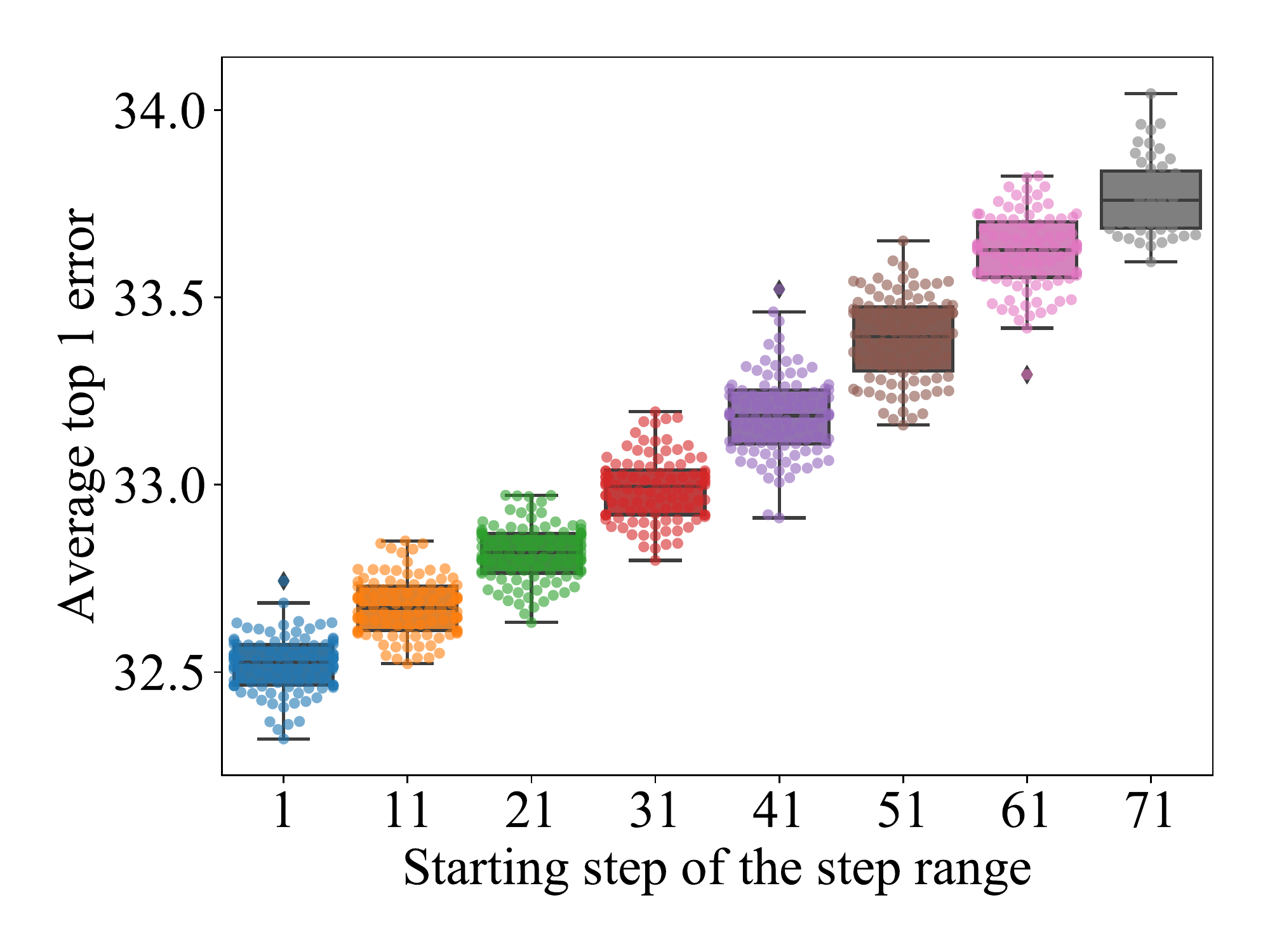}
    \caption{Statistical result of rewiring path}
    \label{fig:139_stat}
\end{figure}

We choose the graph in inferior rewiring as the starting point and repeat our rewiring strategy for 100 times. We use 100 random seeds to generate different random rewiring processes. Figure \ref{fig:139_stat} shows the statistical result of rewiring result. We select continuous step periods, e.g., step 1 to step 10 as one step range and then step 11 to step 20 as another step range. Then we calculate the average of real MLP top 1 errors of each step range. As the number of rewiring steps increases, the median number of average top 1 error increases. This trend is consistent with our prediction, i.e., the statistical increase in top 1 error has a same trend as the predicted increase in top 1 error. Hence, the performance predictor $F(G;\bm\theta)$ can be used for predicting MLP performance.

\section{Discussion}

The performance of neural network architecture can be related to properties of corresponding relational graph. For example, the average path length is a measurement of efficiency of information transport over a graph. At the same time, it is a good indicator of MLP performance. In general, a lower average path length is preferred over a higher one (refer to \ref{appendix:all_features}). However, there is no continuous function fitting well a single graph property and MLP performance. It is impractical to just rely on one graph feature to predict MLP performance. A combination of graph properties, nevertheless, can give a reasonably good prediction of the performance. By predicting a learning error, we are able to search an optimal neural network architecture just by rewiring a graph and calculating graph properties. The benefit of avoiding performing machine learning over an entire graph space is high time efficiency and low computational cost. We adopt linear regression to calculate predicted performance based on graph properties, this process is computationally cheap. Because the graph space is unbounded and does not rely on typical manual designed architecture, the search space is unbiased. Overal, the real MLP performance is consistent with the predicted MLP performance. Based on predicted MLP performance, we are able to target inferior or superior graph structure.

We find different properties might have a similar effect in machine learning accuracy. In other words, some properties are interchangeable. The interchangeability is consistent to physical meaning of those properties. For instance, the average path length, diameter and eccentricity are all related to message exchange efficiency. They are interchangeable in terms of predicting machine learning performance. A directed rewiring strategy, targeting at changing a specific group of graph properties, will definitely enhance the search efficiency.

\section*{Acknowledgments}

This work is supported by Center for Computational Innovation (CCI) of Rensselaer Polytechnic Institute (RPI).

\appendix

\section{Features of a Graph Network} \label{appendix:all_features}

{\noindent \it Average degree.} 
The average degree $\bar{k}$ calculates the average number of edges for one node \cite{luce1949method}. For a graph with $n$ nodes, the average degree $\bar{k}$ is given by:

\begin{equation}
\bar{k} = \frac{\sum_{i \in N}k_{i}}{n},
\end{equation}
where $1 \le i \le n$ and $k_{i}$ is the degree of node $i$. For a complete undirected graph, the average degree is $n - 1$.

{\noindent \it Clustering coefficient.} The clustering coefficient describes the likelihood of a node $j$ in the neighborhood $N_{i}$ of node $i$, is immediately connected to other nodes in $N_{i}$ \cite{watts1998collective}. The clustering coefficient $C_{i}$ for node $i$ in an undirected graph is given by:

\begin{equation}
    C_{i} = \frac{|\{e_{jk}, j \in N_{i}, k \in N_{i}, e_{jk} \in E, j \ne k\}|}{
    \begin{pmatrix}
        k_{i}\\
        2
    \end{pmatrix}
    },
\end{equation}
where $e_{jk}$ is the edge connecting node $j$ and $k$, $E$ is the set of edges for graph $G$. Notation $|\{e_{jk}, j \in N_{i}, k \in N_{i}, e_{jk} \in E, i \ne j\}|$ represents the number of edges within the neighbourhood of node $i$. A larger clustering coefficient physically means nodes in a graph are more likely to cluster together. We use the average clustering coefficient $\frac{1}{n}\sum_{i \in N}C_{i}$ as clustering coefficient $C$ of a graph.

The heterogeneity $\mathcal{H}$ of an unweighted graph is the ratio of variance of degree $k$ to expectation of $k$ and given by \cite{gao2016universal}:

\begin{equation}
    \mathcal{H} = \frac{\frac{1}{n}\sum_{i \in N}k_{i}^{2} - (\frac{1}{n}\sum_{i \in N }k_{i})^{2}}{\frac{1}{n}\sum_{i \in N}k_{i}}
\end{equation}

For a graph with homogeneous degree distribution, heterogeneity is equal to 0.

The average path length $\bar{l}$ measures the average distance between any two nodes in the network and is defined by sum of shortest path length between all pairs of nodes normalized by the total number of node pairs \cite{albert2002statistical,achard2007efficiency}:

\begin{equation}
    \bar{l} = \frac{\sum_{i, j \in N, i \ne j} l_{i, j}}{n (n - 1)}
\end{equation}

Where $l_{i, j}$ is the shortest path length between node $i$ and $j$. The average path length measures the efficiency of message passing over a graph.

The modularity describes the quality of partition of a graph into communities. A good partition separates nodes in such way that majority of edges is in communities and minority lies between them \cite{clauset2004finding,newman2004finding}. The modularity $Q$ is defined by:

\begin{equation}
    Q = \frac{1}{2n}\sum_{i, j \in N}(A_{ij} - \frac{k_{i}k_{j}}{2n})\delta(c_{i}, c_{j})
\end{equation}

Where $A$ is the adjacency matrix of a graph $G$, the $\delta$-function $\delta(c_{i}, c_{j})$ is 1 if $i$ and $j$ are in the same community 0 otherwise. Bimodularity $Q_{b}$ measures the quality of partitioning a graph into two blocks using the Kernighan-Lin algorithm. Kernighan-Lin algorithm partitions nodes of a graph in the manner of minimizing the costs on cutting edges. Greedy modularity $Q_{g}$ measures the quality of partitioning nodes using Clauset-Newman-Moore greedy modularity maximization \cite{clauset2004finding}.

The resilience parameter $\beta_{\text{eff}}$ of an undirected graph is given by \cite{gao2016universal}:

\begin{equation}
    \beta_{\text{eff}} = \frac{\frac{1}{n}\sum_{i \in N}k_{i}^{2}}{\frac{1}{n}\sum_{i \in N}k_{i}}
\end{equation}

The physical meaning of resilience is the ability of a system to maintain basic functionality after external perturbation.

The degree entropy $H$, a measure of disorder, calculates the entropy of the degree distribution and is given by \cite{ji2021generating}:

\begin{equation}
    H = \frac{1}{n}\sum_{i \in N} - \frac{k_{i}}{m} \log\frac{k_{i}}{m}
\end{equation}

Where $m$ is the total number of edges.

The wedge count $W$ counts the number of wedge that is defined as a two-hop path in an undirected graph. $W$ can be calculated by \cite{ji2021generating,gupta2016decompositions}:

\begin{equation}
    W = \sum_{i \in N} 
        \begin{pmatrix}
            k_{i}\\
            2
        \end{pmatrix}
\end{equation}

The Gini index $\mathcal{G}$ measures sparsity of a graph and is defined by \cite{goswami2018sparsity,ji2021generating}:

\begin{equation}
    \mathcal{G} = \frac{2\sum_{i \in N} i\hat{k}_{i}}{n\sum_{i \in N}\hat{k}_{i}} - \frac{n + 1}{n}
\end{equation}

Where $\hat{k}_{i}$ is the degree of node i after sorting degrees.

The average node connectivity $\bar{\kappa}$ is the average of local node connectivity over all pairs of nodes of a graph $G$ and defined as \cite{beineke2002average}:

\begin{equation}
    \bar{\kappa} = \frac{\sum_{i, j \in N, i \ne j} \kappa(i, j)}{
        \begin{pmatrix}
            n\\
            2
        \end{pmatrix}}
\end{equation}

Where $\kappa(i, j)$ is defined as the maximum value of $\kappa$ for which node $i$ and $j$ are $\kappa$-connected. For node $i$ and $j$ considered as $\kappa$-connected, there are $\kappa$ or more pairwise internally disjoint paths between them.

The edge connectivity is the minimum number of edges needed in order to disconnect a graph $G$ \cite{esfahanian2013connectivity}.

The closeness centrality $\mathcal{C}_{i}$ for a node $i$ is the reciprocal of the average shortest path length $l$ of node $i$ to all other $n_{r} - 1$ nodes \cite{freeman1978centrality}:

\begin{equation}
    \mathcal{C}_{i} = \frac{n_{r} - 1}{\sum_{j \in N, j \ne i}l_{i, j}}
\end{equation}

The improved closeness centrality $\mathcal{C}_{i}^{WF}$ by Wasserman and Faust adds a scale factor to scale down closeness centrality for unconnected graph and is given by \cite{wasserman1994social}: 

\begin{equation}
    \mathcal{C}_{i}^{WF} = \frac{n_{r} - 1}{n - 1}\frac{n_{r} - 1}{\sum_{j \in N, j \ne i}l_{i, j}}
\end{equation}

For a connected graph, there is no difference between the close centrality $\mathcal{C}_{i}$ and the improved closeness centrality $\mathcal{C}_{i}^{WF}$.

The eccentricity is defined as the maximum distance from node in a graph to all other nodes \cite{hage1995eccentricity}. The diameter of a graph is the maximum eccentricity while the radius of a graph is the minimum eccentricity.

The average edge betweenness centrality calculates the average of betweenness centrality $\mathcal{C}_{B}(e_{k})$ for all edges of a graph. The betweenness centrality for an edge $e_{k}$ is the sum of the fraction of all shortest paths passing through $e_{k}$. The betweenness centrality can be calculated by \cite{brandes2001faster,brandes2008variants}:

\begin{equation}
    \mathcal{C}_{B}(e_{k}) = \sum_{i, j \in N, i \ne j} \frac{\sigma_{i, j}\bigr\rvert_{e_{k}}}{\sigma_{i, j}}
\end{equation}

Where $\sigma_{i, j}$ is the number of shortest paths connecting node $i$ and $j$. $\sigma_{i, j}\bigr\rvert_{e_{k}}$ is the number of those paths passing through edge $e_{k}$.

Similar to the average edge betweeness centrality, the average node betweenness centrality calculates the average of betweenness centrality $\mathcal{C}_{B}(n)$ for all nodes of a graph. $\mathcal{C}_{B}(n)$ is expressed as:

\begin{equation}
    \mathcal{C}_{B}(k) = \sum_{i, j \in N, i \ne j} \frac{\sigma_{i, j}\bigr\rvert_{k}}{\sigma_{i, j}}
\end{equation}

Where $\sigma_{i, j}\bigr\rvert_{k}$ is the number of shortest paths passing through node $k$.

The central point of dominance $\mathcal{C}_{B}'$ is expressed as \cite{brandes2008variants}:

\begin{equation}
    \mathcal{C}_{B}' = \frac{\sum_{i \in N} (\max_{i \in N} {C}_{B}(i) - {C}_{B}(i))}{n - 1}
\end{equation}

The core number of a node is the largest value $k$ for a $k$-core subgraph containing nodes of degree larger or equal to $k$.

The minimal Laplacian spectrum is the minimum and maximum eigenvalue of the Laplacian matrix of a graph $G$.

The transitivity is defined as the ratio of the number of triangles to the number of triads that are consisted of two edges with a shared node.

The efficiency of a pair of nodes is the inverse of the shortest path length between these two nodes. The local efficiency $\mathcal{E}_{L}(i)$ is the average efficiency of the neighbors $N_{i}$ of node $i$ and defined as \cite{latora2001efficient}:

\begin{equation}
    \mathcal{E}_{L}(i) = \frac{1}{n(n - 1)}\sum_{i, j \in N_{i}, i \ne j}\frac{1}{l_{i, j}}
\end{equation}

We use the average local efficiency of all nodes in a graph $G$ as the local efficiency of $G$.

The global efficiency is the average efficiency of all pairs of nodes.

\section{MLP performance and Features} \label{appendix:features_plot}

\begin{figure*}[hbt]
    \centering
    \includegraphics[width=0.8\textwidth]{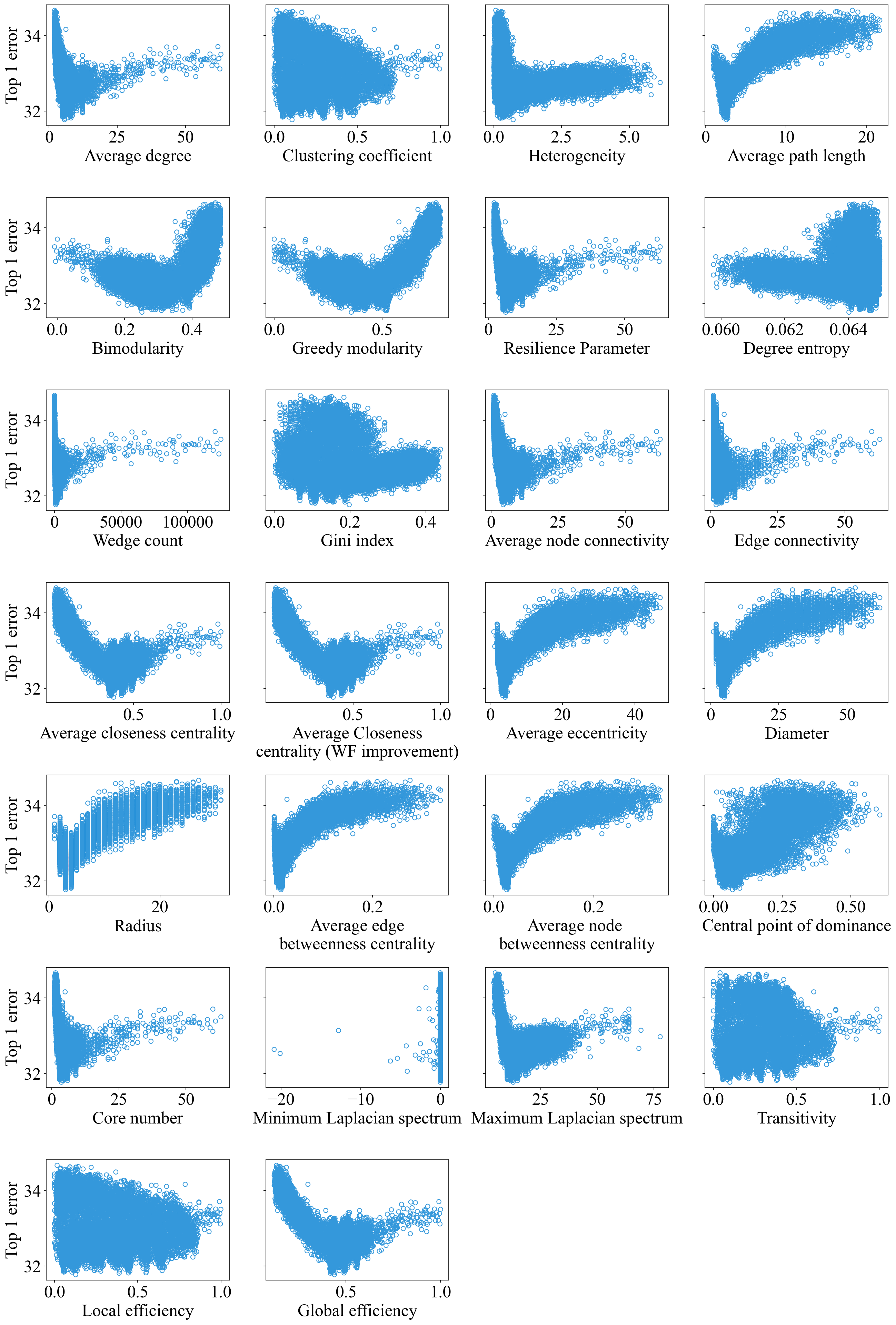}
    \caption{Top 1 errors as a function of different features. Total number of graphs is $19724$}
    \label{fig:mlp_and_features}
\end{figure*}

Figure \ref{fig:mlp_and_features} shows the projection of MLP top 1 error into each feature space. A good fitting of single feature and top 1 error normally requires a discontinuous function. But after a linear combination of those features can give a good fitting quliaty.

\section{SFS after Fixing First Feature} \label{appendix:feature_fix_plot}

\begin{figure*}
    \centering
    \includegraphics[width=0.7\textwidth]{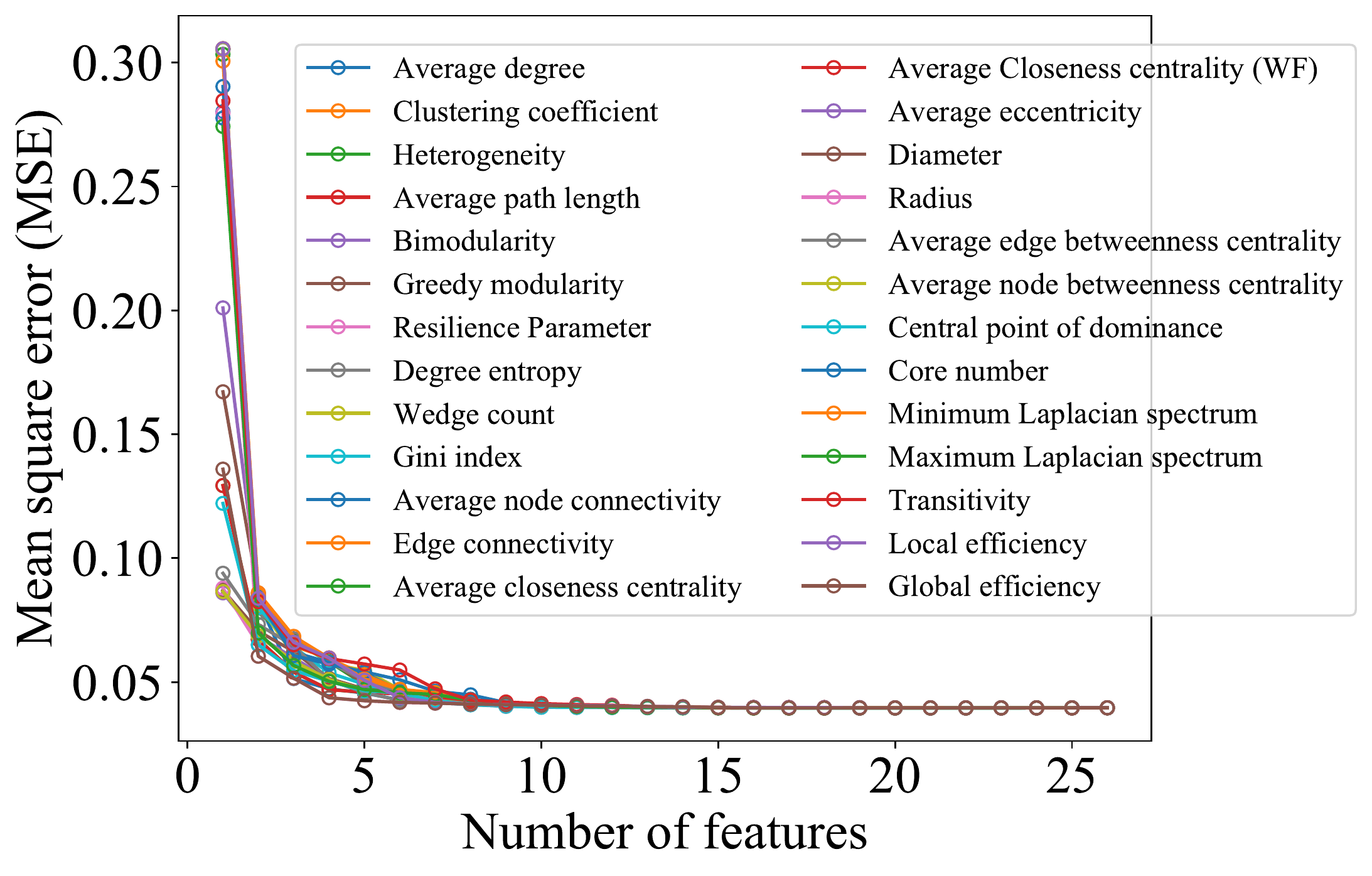}
    \caption{MSE variation with adding features}
    \label{fig:mse_add_ft}
\end{figure*}

\begin{figure*}
    \centering
    \includegraphics[width=0.7\textwidth]{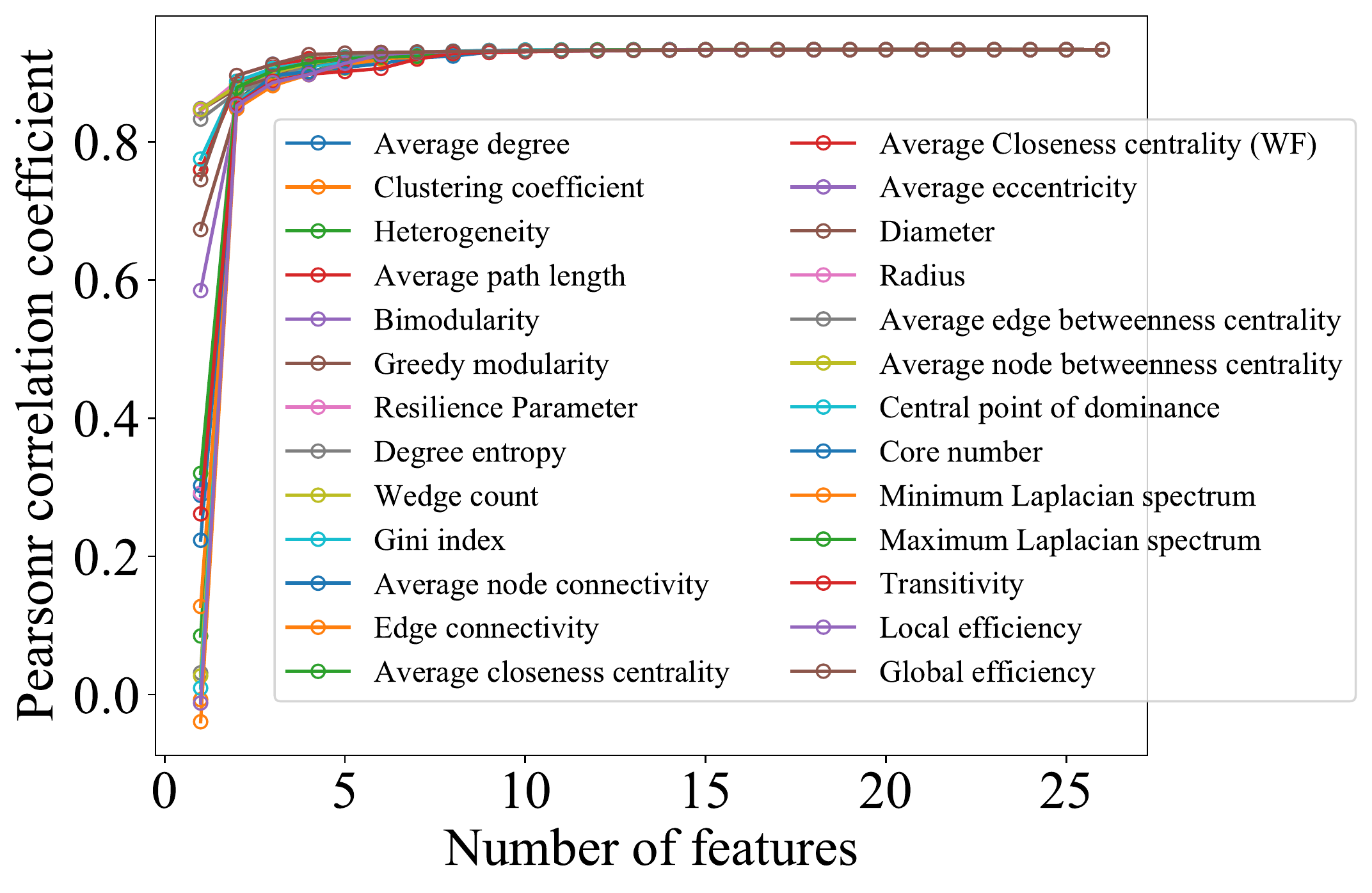}
    \caption{Pearson correlation coefficient variation with adding features}
    \label{fig:pr_add_ft}
\end{figure*}

Figure \ref{fig:mse_add_ft} shows the MSE variation as adding features. Figure \ref{fig:pr_add_ft} shows Pearson correlation coefficient as adding features. Despite the difference in the first feature, they all show a similar trend with adding more features.

\bibliographystyle{named}
\bibliography{ijcai22}

\end{document}